\documentclass[sigconf]{acmart}
\setcopyright{none}
\AtBeginDocument{%
  }

\begin{document}

\title{Compression as an Adversarial Amplifier \\ Through Decision Space Reduction}

\author{Lewis Evans}
\authornote{Both authors contributed equally to this research.}
\author{Harkrishan Jandu}
\authornotemark[1]
\affiliation{%
  \institution{University of Nottingham}
  \city{Nottingham}
  \country{United Kingdom}
}

\author{Zihan Ye}
\affiliation{%
  \institution{University of Chinese Academy of Sciences}
  \city{Beijing}
  \country{China}}

\author{Yang Lu}
\affiliation{%
  \institution{Xiamen University}
  \city{Xiamen}
  \country{China}
}

\author{Shreyank N Gowda}
\affiliation{%
 \institution{School of Computer Science, University of Nottingham}
 \city{Nottingham}
 \country{United Kingdom}
 }

\renewcommand{\shortauthors}{L. Evans et al.}

\begin{abstract}
Image compression is a ubiquitous component of modern visual pipelines, routinely applied by social media platforms and resource-constrained systems prior to inference. Despite its prevalence, the impact of compression on adversarial robustness remains poorly understood. We study a previously unexplored adversarial setting in which attacks are applied directly in compressed representations, and show that compression can act as an adversarial amplifier for deep image classifiers. Under identical nominal perturbation budgets, compression-aware attacks are substantially more effective than their pixel-space counterparts. We attribute this effect to decision space reduction, whereby compression induces a non-invertible, information-losing transformation that contracts classification margins and increases sensitivity to perturbations. Extensive experiments across standard benchmarks and architectures support our analysis and reveal a critical vulnerability in compression-in-the-loop deployment settings. Code will be released.
\end{abstract}



\maketitle

\section{Introduction}

Over the past decade, deep learning has undergone rapid and transformative progress, achieving remarkable performance across vision, language, and multimodal tasks~\cite{deng2009imagenet,krizhevsky2012imagenet,gowda2021smart,feichtenhofer2019slowfast}. Modern models now rival or surpass human-level accuracy in many benchmarks, driven by advances in large-scale training, data availability, and architectural innovations~\cite{gpt,bai2023qwen,gemini}. However, this progress has been accompanied by a growing fragility: these systems are increasingly vulnerable to adversarial manipulation. In particular, recent work has highlighted how easily state-of-the-art models can be ``jailbroken'' through carefully crafted prompts or inputs~\cite{jailbreak1,jailbreak2,jailbreak3,jailbreak4,jailbreak5}, exposing unintended behaviours and bypassing safety constraints. Similarly, adversarial attacks~\cite{alam2025adversarial,croce2020robustbench} that are often imperceptible to humans can reliably induce incorrect or even harmful outputs. The ease with which such vulnerabilities can be exploited raises serious concerns about the robustness and trustworthiness of deployed systems, especially as they are integrated into real-world, safety-critical applications.

Image compression is a foundational component of modern visual data pipelines~\cite{laghari2018assessment,chen2025information}. Images processed by social media platforms, messaging services, content delivery networks, and edge devices are almost invariably compressed prior to storage, transmission, or inference~\cite{chlubna2025comparative,kohne2022practical}. Compression enables significant reductions in bandwidth, memory, and latency, and is therefore tightly coupled to real-world deployment. Despite this, adversarial robustness is still predominantly studied under the assumption that models operate directly on uncompressed pixel representations~\cite{chakraborty2021survey,wei2024physical}.

Existing work has often characterized compression as a defensive or purifying operation~\cite{jia2019comdefend,das2018compression,ferrari2023compress}. Prior studies have shown that compression can suppress high-frequency perturbations and partially mitigate adversarial attacks crafted in pixel space~\cite{das2018compression,ferrari2023compress}. This perspective has contributed to the widespread belief that compression improves robustness or, at minimum, does not increase vulnerability. However, this view implicitly assumes a threat model in which adversarial perturbations are applied before compression and overlooks settings in which attacks may be applied after compression or directly within compressed representations.

In this work, we study a compression-aware adversarial setting that more closely reflects practical deployment pipelines. We show that compression can act as an adversarial amplifier, substantially increasing the effectiveness of adversarial perturbations under fixed nominal budgets. Across datasets, architectures, and compression strengths, attacks applied in compressed representations consistently cause larger output shifts than their pixel-space counterparts.

\begin{figure}[t]
    \centering
    \includegraphics[width=\linewidth]{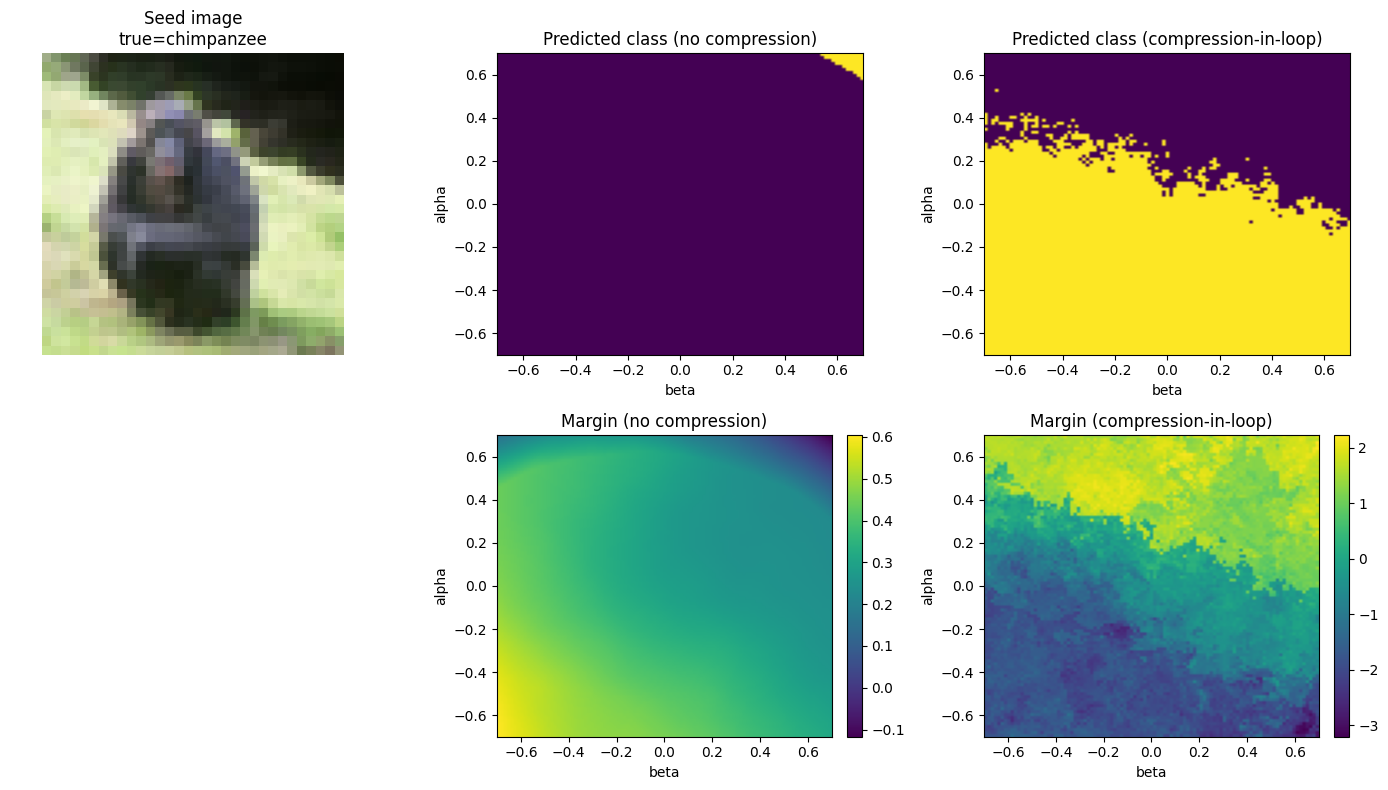}
    \caption{Local decision regions around a correctly classified input are visualized in a 2D neighborhood. Without compression (middle), the true-class region is large and margins are smooth. With compression-in-the-loop (right), the region contracts, competing classes appear closer to the input, and margins collapse, illustrating how compression brings decision boundaries nearer and amplifies vulnerability.}
    \label{fig:teaser}
\end{figure}

We attribute this phenomenon to a geometric mechanism that we term \emph{decision space reduction}. Compression induces a non-invertible, information-losing transformation that contracts the effective decision space around an input. This effect is illustrated in Figure~\ref{fig:teaser}, which shows that compression shrinks the region assigned to the true class and brings decision boundaries closer in the local input neighborhood. This contraction reduces classification margins and increases boundary proximity for compressed representations. As a consequence, perturbations of the same magnitude are more likely to cross decision boundaries in compressed space than in the original image domain. Importantly, this effect is local and geometric rather than tied to a specific attack algorithm. Interestingly, the same mechanism also helps explain why compression-based purification defenses can be effective when applied after perturbation, as compression may suppress boundary-crossing components of adversarial noise and partially restore margins relative to the perturbed input.

We support this analysis through a combination of empirical evaluation and decision space visualization. By examining local neighborhoods around fixed inputs, we show that compression consistently shrinks the region assigned to the true class and increases boundary density across random and gradient-informed directions. Our findings reveal a previously overlooked vulnerability arising from the ubiquitous use of compression and highlight the need to reconsider adversarial robustness under compression-in-the-loop threat models.

This work is guided by the following research questions:
\begin{enumerate}
    \item How does image compression alter the local decision geometry of deep image classifiers?
    \item Does compression systematically reduce classification margins and contract the effective decision space around inputs?
    \item Under fixed perturbation budgets, are adversarial perturbations inherently more damaging in compressed representations than in pixel space?
\end{enumerate}

This paper makes three primary contributions. First, we formalize a compression-aware adversarial setting that reflects realistic deployment pipelines in which inputs are compressed prior to inference. Second, we identify decision space reduction as a fundamental geometric mechanism through which compression amplifies adversarial vulnerability. Third, we provide empirical and visual evidence demonstrating consistent margin collapse and decision boundary contraction induced by compression across standard benchmarks and architectures.

\section{Related Work}

\paragraph{Adversarial robustness.}
Adversarial examples expose vulnerabilities in deep networks where imperceptible perturbations lead to incorrect predictions \cite{chakraborty2021survey, wei2024physical}. Early work focused on constructing attacks such as FGSM~\cite{fgsm} and CW~\cite{cw}, and on defending through adversarial training and certified methods \cite{madry2017towards, tramer2017ensemble, wong2020fast}. Other research has investigated geometric and manifold perspectives, showing that adversarial vulnerability arises from boundary proximity and high codimension of data manifolds \cite{khoury2018geometry, olivier2023many}. Unlike these studies which assume pixel-space perturbations, we examine how compression transforms affect decision boundary geometry.

\paragraph{Input transformations and compression defenses.}
Input transformations including JPEG compression~\cite{wallace1991jpeg} and bit-depth reduction have long been studied as means to mitigate adversarial noise \cite{guo2018countering, das2018compression, jia2019comdefend}. Work such as feature distillation refines JPEG quantization to improve defense efficacy \cite{liu2019feature}, and recent methods propose learned compression to defend against adversarial examples \cite{bell2025persistent}. Evaluations, however, have shown that many simple transformation defenses can be circumvented or are limited without considering adaptive attacks \cite{mahmood2021beware}. These works treat compression as a defensive preprocessing step; our setting instead considers attacks applied after compression and reveals that compression can amplify vulnerability.

\paragraph{Compression in vision and learning.}
Image compression is essential in practical visual systems to reduce bandwidth and storage costs, playing a central role in social media and cloud platforms \cite{laghari2018assessment, chlubna2025comparative, chen2025information}, and its interaction with learning systems has been explored in the context of performance and efficiency \cite{chen2025information}. Recent work also analyzes robustness of learned image compression models themselves under adversarial perturbations \cite{song2024training,sui2024transferable}. These studies primarily focus on compression models as objects of defense; our work instead investigates the \emph{effect of compression on classifier decision geometry} and as an adversarial amplifier.

\paragraph{Decision boundary geometry and robustness.}
The geometry of decision boundaries and classification margins has been linked to robustness, leading to theoretical frameworks that relate curvature, dimensionality, and nearest-boundary distances to vulnerability \cite{fawzi2018analysis, trillos2022adversarial}. Sampling-based metrics have also been proposed to understand boundary persistence and sensitivity \cite{bell2025persistent}. While these works analyze geometric aspects of adversarial susceptibility, they do not address how upstream transformations such as compression systematically reshape the decision space. Our work connects compression to margin collapse and boundary densification in local neighborhoods.

\paragraph{Attacks in transformed and representation spaces.}
Beyond pixel-space perturbations, several works have studied adversarial attacks and robustness in alternative representation domains. These include attacks in feature space, frequency space, and learned latent spaces, which demonstrate that model vulnerability depends strongly on the representation in which perturbations are applied \cite{yin2019fourier, zhaobridging}. Such studies highlight that changing the input representation can alter the effective geometry of adversarial perturbations. However, these methods typically rely on learned or differentiable transforms and focus on manipulating internal or semantic representations. In contrast, we study a \emph{structural compression transform} that is non-invertible and information-reducing, and show that this class of transformations systematically contracts the decision space and reduces margins, independent of any learned latent structure.

\section{Method}

\subsection{Preliminaries and Notation}

Let $f:\mathbb{R}^d \rightarrow \mathbb{R}^K$ denote a $K$-class classifier mapping an input image $x \in \mathbb{R}^d$ to class logits. The predicted label is
\[
\hat{y}(x) = \arg\max_{k} f_k(x).
\]
For a ground-truth class $y$, we define the \emph{classification margin}
\[
m(x) = f_y(x) - \max_{k \neq y} f_k(x).
\]
Standard adversarial robustness considers perturbations applied directly in pixel space:
\[
x' = x + \delta, \quad \|\delta\|_p \le \epsilon,
\]
and studies the stability of $\hat{y}(x')$ under bounded $\delta$.

\subsection{Compression-Aware Inference Model}

We model image compression as a deterministic transformation
\[
C: \mathbb{R}^d \rightarrow \mathcal{Z},
\]
where $\mathcal{Z}$ denotes a compressed representation space. Compression is typically non-invertible, information-reducing, and involves quantization or discretization. In deployment settings, inference may be performed on compressed inputs or representations derived from them. We therefore consider the effective model
\[
g(x) = f(C(x)),
\]
where $C$ defines the input representation seen by the classifier. Unlike stochastic noise or data augmentation, $C$ is a structural transformation of the input domain. Such compression-in-the-loop pipelines are standard in real-world systems, where images transmitted through social media platforms, messaging applications, and cloud services are routinely compressed prior to storage, transmission, or inference \cite{laghari2018assessment, chlubna2025comparative, chen2025information}.

\begin{figure*}[t]
\centering
\includegraphics[width=\linewidth]{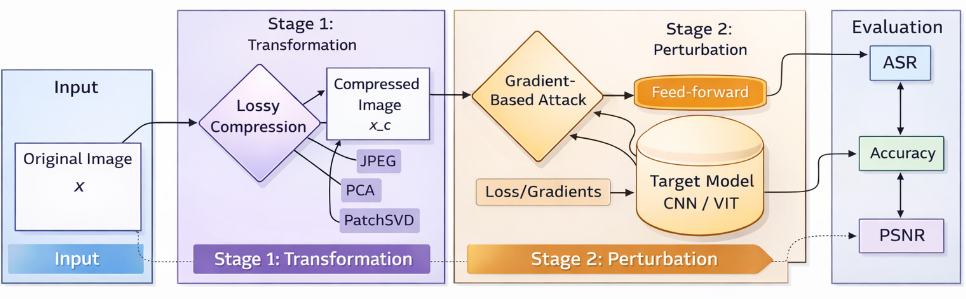}
\caption{Overview of the compression-aware adversarial pipeline. An input image is first transformed by a lossy compression operator, after which a gradient-based adversary perturbs the compressed representation. The perturbed input is then fed to the target model for prediction, and performance is evaluated using accuracy, attack success rate, and distortion metrics such as PSNR.}
\label{fig:overview}
\end{figure*}

An overview of the full pipeline, including compression, perturbation, and evaluation stages, is shown in Figure~\ref{fig:overview}.

\subsection{Threat Model: Compression-Aware Adversary}

In contrast to the standard pixel-space adversarial setting, we consider an adversary that operates after compression. Let
\[
z = C(x)
\]
be the compressed representation of the input. The adversary perturbs $z$ to obtain
\[
z' = z + \eta, \quad \|\eta\|_p \le \epsilon,
\]
and the model prediction becomes $\hat{y} = f(z')$. This defines a \emph{compression-aware} adversarial setting in which the perturbation budget is measured in the compressed representation space rather than the original pixel domain. This differs from purification-based defenses, which assume perturbations are applied before compression.

\paragraph{Implementation of Compression-Aware Perturbations.}
A key implementation detail is how gradients are computed in the presence of non-differentiable compression operators such as JPEG. In our setting, we do not backpropagate through the compression operator $C$. Instead, we first compute the compressed representation $z = C(x)$ using a standard non-differentiable implementation (e.g., libjpeg), and treat $z$ as the input to the model. Adversarial perturbations are then applied directly to $z$, i.e., $z' = z + \eta$, with gradients computed with respect to $z$ through the classifier $f$.

This corresponds to a threat model in which the attacker operates \emph{after} compression, and therefore does not require gradients through $C$. In particular, we do not use differentiable JPEG approximations or straight-through estimators. This design choice reflects practical deployment settings where compression is a fixed preprocessing step and not part of the differentiable computation graph. We note that this setting differs from prior work on compression-based defenses, which often assume perturbations are applied before compression and require gradient approximations through $C$.

\subsection{Decision Space Reduction}

We characterize the effect of compression through local decision geometry. Let $\mathcal{N}(x)$ denote a neighborhood around an input $x$. The \emph{true-class decision region} is
\[
\mathcal{R}(x) = \{x' \in \mathcal{N}(x) : \hat{y}(x') = y\}.
\]
Under compression-aware inference, this becomes
\[
\mathcal{R}_C(x) = \{x' \in \mathcal{N}(x) : f(C(x')) = y\}.
\]
We define \emph{decision space reduction} as the contraction of the true-class region under compression:
\[
\mathrm{Vol}(\mathcal{R}_C(x)) < \mathrm{Vol}(\mathcal{R}(x)),
\]
where $\mathrm{Vol}(\cdot)$ denotes a measure over $\mathcal{N}(x)$. Intuitively, compression reduces the volume of perturbations that preserve the correct class, bringing decision boundaries closer to the input.

\subsection{Theoretical Insight: Margin Shrinkage Under Compression}

We provide a simple geometric observation linking compression to reduced robustness.

\textbf{Proposition 1.}  
Let $C:\mathbb{R}^d \rightarrow \mathcal{Z}$ be a non-invertible compression operator and $g(x)=f(C(x))$. Assume $C$ is locally Lipschitz with constant $L_C$ and $f$ is locally Lipschitz with constant $L_f$. Then the effective local robust radius of $g$ around $x$ satisfies
\[
r_g(x) \le \frac{m(C(x))}{L_f L_C}.
\]

\textit{Proof sketch.}
Under first-order approximation, a perturbation $\delta$ in input space produces a change
\[
\Delta f \approx \nabla_z f(C(x))^\top (C(x+\delta) - C(x)).
\]
Using Lipschitz continuity,
\[
\|C(x+\delta) - C(x)\| \le L_C \|\delta\|.
\]
Thus the change in logits is bounded by $L_f L_C \|\delta\|$. The decision boundary is crossed when this exceeds the margin $m(C(x))$, yielding the stated bound.

\subsection{Sensitivity Amplification}

Local robustness can be approximated by first-order analysis. For a small perturbation $\eta$ in compressed space,
\[
\Delta f \approx \nabla_z f(z)^\top \eta.
\]
A common proxy for the local robust radius is
\[
r(x) \approx \frac{m(x)}{\|\nabla f(x)\|}.
\]
Compression can reduce the margin $m(C(x))$ and alter gradient magnitudes, effectively shrinking $r(x)$. Consequently, perturbations of the same norm in compressed space are more likely to change the model output, leading to \emph{sensitivity amplification}.

\subsection{Quantifying Decision Space Reduction}

To empirically characterize decision space reduction, we evaluate local neighborhoods around inputs using the following metrics:

\begin{itemize}
    \item \textbf{True-class region fraction:} proportion of sampled neighborhood points classified as the true class.
    \item \textbf{Mean margin:} average value of $m(x')$ over the neighborhood.
    \item \textbf{Negative-margin fraction:} proportion of points where $m(x') < 0$.
    \item \textbf{Boundary density:} frequency of class transitions within the neighborhood.
\end{itemize}

These quantities provide measurable proxies for region contraction and boundary proximity.

\subsection{Relation to Purification Approaches}

Purification-based defenses assume perturbations are applied in pixel space and that compression removes adversarial noise, modeling inputs as $C(x + \delta)$. In contrast, our setting considers perturbations applied after compression, $C(x) + \eta$. This shift in threat model exposes compression as a potential attack surface rather than solely a defense mechanism.

\subsection{Decision Space Reduction as a Unifying View of Amplification and Purification}

The decision space reduction perspective also provides insight into why compression-based \emph{purification} defenses can be effective when compression is applied \emph{after} an adversarial perturbation. Consider a perturbed input $x' = x + \delta$ that lies near or across a decision boundary in pixel space. Applying compression yields $C(x')$, which can be viewed as a projection onto a lower-dimensional, quantized manifold. High-frequency or low-energy components of $\delta$ that were responsible for crossing the boundary may be attenuated or removed, effectively moving the representation back toward the interior of the true-class region in compressed space.

Geometrically, this corresponds to a partial \emph{re-expansion} of the effective decision region relative to $x'$. While compression of a clean input $x$ contracts the region $\mathcal{R}(x)$ to $\mathcal{R}_C(x)$, compression of an already perturbed input can reduce the effective displacement from the decision boundary. Thus, the same transformation $C$ can either contract or effectively restore margins depending on whether it is applied before or after perturbation.

This asymmetry explains the strong order effects observed in Section~4.5. When compression precedes the attack ($C(x) + \eta$), the true-class region is already reduced, making it easier for small perturbations to cross the boundary. When compression follows the attack ($C(x + \delta)$), components of $\delta$ that caused boundary crossing may be suppressed, increasing the margin relative to $x'$. Decision space reduction therefore unifies both adversarial amplification and purification within a single geometric framework.

\section{Experiments}

\subsection{Setup}

We evaluate compression-aware adversarial vulnerability on CIFAR-10, CIFAR-100, and ImageNet (validation set). Models include standard ResNet architectures of varying depth. We compare conventional pixel-space attacks against compression-aware variants in which inputs are first compressed and perturbations are applied in the compressed domain. We report classification accuracy under attack and Peak Signal-to-Noise Ratio (PSNR) to quantify perceptual distortion.

Attacks include FGSM~\cite{fgsm}, PGD~\cite{madry2017towards}, AutoAttack~\cite{croce2020reliable}, BSR~\cite{bsr}, and Attention-Fool~\cite{alam2025adversarial}. Compression-aware variants apply JPEG compression~\cite{wallace1991jpeg}, PCA compression~\cite{pca} and PatchSVD~\cite{patchsvd} prior to perturbation. Perturbation budgets are identical across settings to isolate the effect of compression on robustness. We also want to add that whilst the PSNR is relatively high for AutoAttack, the time needed to run it is also significantly higher. AutoAttack~\cite{croce2020reliable} has been estimated to take 5-10 times the overall time of PGD for example. This is expected as AutoAttack is essentially an ensemble attack and used to report worst-case robust performance. 

\subsection{Implementation Details}
To measure attack performance across different model architectures, we experimented with the following models: ResNet-18, ResNet-34 (results in appendix), ResNet-50 (results in appendix), ViT-B/16, ViT-B/32 (results in appendix), and ViT-L/16. All architectures were initialized with pretrained ImageNet-1K weights provided by the Torchvision model repository. No additional fine-tuning was performed. 

All experiments were executed on a compute server equipped with NVIDIA RTX A4000 GPUs, with one GPU allocated per model. In addition, 8 logical CPU cores and 24\,GB of system memory were used.

All decision-space visualizations use 2D planes constructed from the loss gradient direction and a random orthogonal direction. Grids use $61\times61$ samples over a radius of $0.35$. JPEG compression uses libjpeg with standard quality parameters. Gradients are computed with frozen BatchNorm statistics. Random seeds are fixed for reproducibility.

\begin{table}[t]
\centering
\caption{Attacks and default hyperparameters.}
\label{tab:attack-hparams}
\resizebox{\linewidth}{!}{%
\begin{tabular}{l l}
\hline
Attack & Default setting \\
\hline
FGSM & $\epsilon = 0.01$ \\
PGD (L$_\infty$) & $\epsilon_{\text{pixel}}=2/255$, $\alpha_{\text{pixel}}=1/255$, iters$=5$, random\_start=True \\
JPEG & quality $=25$ \\
JPEG $\rightarrow$ FGSM & quality $=55$, $\epsilon=0.02$ \\
JPEG $\rightarrow$ PGD & quality $=55$, $\epsilon_{\text{pixel}}=0.02$, iters$=10$ \\
PCA & n\_components $=22$ \\
PCA $\rightarrow$ FGSM & n\_components $=50$, $\epsilon_{\text{pixel}}=0.02$ \\
PCA $\rightarrow$ PGD & n\_components $=50$, $\epsilon=0.02$, iters$=10$ \\
\hline
\end{tabular}%
}
\end{table}

\subsection{Main Robustness Results}

\begin{table}[t]
\centering
\caption{Robustness on CIFAR-10 under compression-aware attacks. All our attacks are in \textcolor{blue}{blue} and the best performer is \textbf{bolded}.}
\label{tab:cifar10}
\resizebox{\linewidth}{!}{%
\begin{tabular}{l l c c}
\hline
Model & Attack & Accuracy (\%) & PSNR \\
\hline

ResNet-18 & Clean & 94.98 & 100.00 \\
& FGSM ($\epsilon=0.06$) & 18.72 & 24.18 \\

& BSR ($\epsilon=0.06$) & 87.70 & 39.77 \\

& AutoAttack & 1.32 & 32.24 \\
& PGD ($\epsilon=0.06$) & 0.00 & 28.68 \\
& Attention-Fool & 28.85 & 29.34 \\
& \textcolor{blue}{JPEG $\rightarrow$ FGSM} & \textcolor{blue}{13.14} & \textcolor{blue}{29.66} \\
& \textcolor{blue}{PCA $\rightarrow$ FGSM} & \textcolor{blue}{18.11} & \textcolor{blue}{24.08} \\
& \textcolor{blue}{PatchSVD $\rightarrow$ FGSM} & \textcolor{blue}{17.21} & \textcolor{blue}{24.04} \\
& \textcolor{blue}{JPEG $\rightarrow$ PGD} & \textbf{\textcolor{blue}{0.00}}& \textcolor{blue}{25.21}\\
& \textcolor{blue}{PCA $\rightarrow$ PGD} & \textcolor{blue}{0.00} & \textcolor{blue}{28.47} \\
& \textcolor{blue}{PatchSVD $\rightarrow$ PGD} & \textcolor{blue}{0.00} & \textcolor{blue}{28.38} \\
\hline

ResNet-50 & Clean & 94.65 & 100.00 \\
& FGSM ($\epsilon=0.06$) & 15.78 & 24.18 \\
& BSR ($\epsilon=0.06$) & 81.96 & 39.74 \\
& PGD ($\epsilon=0.06$) & 0.00 & 26.90\\
& AutoAttack & 1.56 & 32.24 \\
& Attention-Fool & 31.47 & 29.64 \\
& \textcolor{blue}{JPEG $\rightarrow$ FGSM} & \textcolor{blue}{15.58} & \textcolor{blue}{29.66} \\
& \textcolor{blue}{PCA $\rightarrow$ FGSM} & \textcolor{blue}{15.20} & \textcolor{blue}{24.07} \\
& \textcolor{blue}{PatchSVD $\rightarrow$ FGSM} & \textcolor{blue}{15.05} & \textcolor{blue}{24.04} \\
& \textcolor{blue}{JPEG $\rightarrow$ PGD} & \textbf{\textcolor{blue}{0.00}}& \textcolor{blue}{25.22}\\
&\textcolor{blue}{ PCA $\rightarrow$ PGD} & \textcolor{blue}{0.00}& \textcolor{blue}{28.65} \\
&\textcolor{blue}{ PatchSVD $\rightarrow$ PGD} & \textcolor{blue}{0.00}& \textcolor{blue}{28.57} \\
\hline
ViT-L-16 & Clean & 98.12 & 100.00 \\
& FGSM ($\epsilon=0.06$) & 22.67 &  25.12\\
& BSR ($\epsilon=0.06$) & 89.63 & 37.75 \\
& PGD ($\epsilon=0.06$) & 12.05 & 28.87  \\
& AutoAttack & 8.74 & 31.64 \\
& Attention-Fool & 29.15 & 29.29 \\
& \textcolor{blue}{JPEG $\rightarrow$ FGSM} & \textcolor{blue}{15.19} & \textcolor{blue}{29.48} \\
& \textcolor{blue}{PCA $\rightarrow$ FGSM} & \textcolor{blue}{15.88} & \textcolor{blue}{24.52} \\
& \textcolor{blue}{PatchSVD $\rightarrow$ FGSM} & \textcolor{blue}{15.56} & \textcolor{blue}{24.55} \\
& \textcolor{blue}{JPEG $\rightarrow$ PGD} & \textbf{\textcolor{blue}{4.89}} & \textcolor{blue}{26.24} \\
&\textcolor{blue}{ PCA $\rightarrow$ PGD} & \textcolor{blue}{11.12} & \textcolor{blue}{28.14} \\
&\textcolor{blue}{ PatchSVD $\rightarrow$ PGD} & \textcolor{blue}{11.36} & \textcolor{blue}{28.17} \\
\hline

\end{tabular}%
}
\end{table}

\begin{table}[t]
\centering
\caption{Robustness on CIFAR-100 under compression-aware attacks. All our attacks are in \textcolor{blue}{blue} and the best performer is \textbf{bolded}.}
\label{tab:cifar100}
\resizebox{\linewidth}{!}{%
\begin{tabular}{l l c c}
\hline
Model & Attack & Accuracy (\%) & PSNR \\
\hline
ResNet-18 & Clean & 79.26 & 100.0 \\
          & FGSM ($\epsilon=0.03$) & 10.03 & 30.22 \\
          & PGD ($\epsilon=0.03$) & 0.01 & 34.15 \\
          & BSR ($\epsilon=0.03$) & 68.44 & 42.65 \\
          & Attention-Fool & 11.84 & 30.82 \\
          
& \textcolor{blue}{JPEG $\rightarrow$ FGSM} & \textcolor{blue}{8.46} & \textcolor{blue}{29.82} \\
& \textcolor{blue}{PCA $\rightarrow$ FGSM} & \textcolor{blue}{9.77} & \textcolor{blue}{29.84} \\
& \textcolor{blue}{PatchSVD $\rightarrow$ FGSM} & \textcolor{blue}{9.42} & \textcolor{blue}{29.71} \\
& \textcolor{blue}{JPEG $\rightarrow$ PGD} & \textcolor{blue}{4.73} & \textcolor{blue}{28.71} \\
&\textcolor{blue}{ PCA $\rightarrow$ PGD} & \textcolor{blue}{\textbf{0.00}} & \textcolor{blue}{33.41} \\
&\textcolor{blue}{ PatchSVD $\rightarrow$ PGD} & \textcolor{blue}{\textbf{0.01}} & \textcolor{blue}{33.16} \\          
\hline
ResNet-50 & Clean & 80.93 & 100.0 \\
          & FGSM ($\epsilon=0.03$) & 17.47 & 30.22 \\
          & PGD ($\epsilon=0.03$) & 0.12 & 34.56 \\
          & JPEG $\rightarrow$ PGD & 8.91 & 28.71 \\
          & BSR ($\epsilon=0.03$) & 70.32 & 42.68 \\
          & AutoAttack & 5.61 & 34.29 \\
          & Attention-Fool & 9.84 & 29.89 \\
& \textcolor{blue}{JPEG $\rightarrow$ FGSM} & \textcolor{blue}{13.94} & \textcolor{blue}{29.83} \\
& \textcolor{blue}{PCA $\rightarrow$ FGSM} & \textcolor{blue}{16.66} & \textcolor{blue}{29.84} \\
& \textcolor{blue}{PatchSVD $\rightarrow$ FGSM} & \textcolor{blue}{16.66} & \textcolor{blue}{29.71} \\
& \textcolor{blue}{JPEG $\rightarrow$ PGD} & \textcolor{blue}{8.91} & \textcolor{blue}{28.71} \\
&\textcolor{blue}{ PCA $\rightarrow$ PGD} & \textcolor{blue}{\textbf{0.07}} & \textcolor{blue}{33.48} \\
&\textcolor{blue}{ PatchSVD $\rightarrow$ PGD} & \textcolor{blue}{\textbf{0.05}} & \textcolor{blue}{33.48} \\
\hline

\hline
ViT-L-16 & Clean & 85.87 & 100.00 \\
& FGSM ($\epsilon=0.06$) & 24.57 &  25.12\\
& BSR ($\epsilon=0.03$) & 71.83 & 42.69  \\
& PGD ($\epsilon=0.06$) & 14.47 & 28.87  \\
& AutoAttack & 6.79 & 32.65 \\
& Attention-Fool & 7.85 & 31.34 \\
& \textcolor{blue}{JPEG $\rightarrow$ FGSM} & \textcolor{blue}{11.19} & \textcolor{blue}{29.88} \\
& \textcolor{blue}{PCA $\rightarrow$ FGSM} & \textcolor{blue}{11.88} & \textcolor{blue}{27.52} \\
& \textcolor{blue}{PatchSVD $\rightarrow$ FGSM} & \textcolor{blue}{12.16} & \textcolor{blue}{27.55} \\
& \textcolor{blue}{JPEG $\rightarrow$ PGD} & \textbf{\textcolor{blue}{3.67}} & \textcolor{blue}{28.84} \\
&\textcolor{blue}{ PCA $\rightarrow$ PGD} & \textcolor{blue}{4.59} & \textcolor{blue}{33.45} \\
&\textcolor{blue}{ PatchSVD $\rightarrow$ PGD} & \textcolor{blue}{5.36} & \textcolor{blue}{33.41} \\
\hline

\hline
\end{tabular}%
}
\end{table}

\begin{table}[t]
\centering
\caption{Robustness on ImageNet under compression-aware attacks. All our attacks are in \textcolor{blue}{blue} and the best performer is \textbf{bolded}.}
\label{tab:imagenet}
\resizebox{0.85\linewidth}{!}{%
\begin{tabular}{l l c c}
\hline
Model & Attack & Accuracy (\%) & PSNR \\
\hline
ResNet-18 & Clean & 67.28 & 100.00 \\
          & FGSM  & 15.45 & 14.53 \\
          & PGD & 0.01 & 38.04 \\
          & JPEG & 54.67 & 29.41 \\
          & PCA  & 41.77 & 27.53 \\
          & AutoAttack & 6.64 & 47.84 \\
          & \textcolor{blue}{JPEG $\rightarrow$ FGSM } & \textcolor{blue}{1.47} & \textcolor{blue}{31.41} \\
          & \textcolor{blue}{JPEG $\rightarrow$ PGD } & \textcolor{blue}{\textbf{0.00}} & \textcolor{blue}{30.48} \\
          & \textcolor{blue}{PCA  $\rightarrow$ FGSM } & \textcolor{blue}{0.64} & \textcolor{blue}{30.86} \\
          & \textcolor{blue}{PCA  $\rightarrow$ PGD  }& \textcolor{blue}{\textbf{0.00}} & \textcolor{blue}{32.48} \\
          
\hline
ResNet-50 & Clean & 80.10 & 100.00 \\
          & FGSM  & 52.46 & 14.26 \\
          & PGD  & 9.67 & 38.25 \\
          & AutoAttack  & 3.82 & 49.02 \\
          & JPEG & 66.29 & 29.22 \\
          & PCA & 54.05 & 27.47 \\
          & \textcolor{blue}{JPEG $\rightarrow$ FGSM } & \textcolor{blue}{32.64} & \textcolor{blue}{31.41} \\
          & \textcolor{blue}{JPEG $\rightarrow$ PGD } & \textcolor{blue}{\textbf{0.98}} & \textcolor{blue}{30.57} \\
          & \textcolor{blue}{PCA  $\rightarrow$ FGSM} & \textcolor{blue}{33.46} & \textcolor{blue}{30.71} \\
          & \textcolor{blue}{PCA  $\rightarrow$ PGD } & \textcolor{blue}{1.00} & \textcolor{blue}{32.61} \\
          
\hline
ViT-B/16  & Clean & 80.00 & 100.00 \\
          & FGSM  & 46.78 & 14.37 \\
          & PGD  & 5.12 & 37.93 \\
          & AutoAttack & 5.85 & 47.80 \\
          & JPEG & 70.66 & 29.27 \\
          & PCA  & 67.29 & 27.68 \\
          & \textcolor{blue}{JPEG $\rightarrow$ FGSM } & \textcolor{blue}{33.49} & \textcolor{blue}{31.56} \\
          & \textcolor{blue}{JPEG $\rightarrow$ PGD } & \textcolor{blue}{0.57} & \textcolor{blue}{30.46} \\
          
          & \textcolor{blue}{PCA  $\rightarrow$ FGSM} & \textcolor{blue}{25.22} & \textcolor{blue}{30.86} \\
          & \textcolor{blue}{PCA  $\rightarrow$ PGD } & \textcolor{blue}{\textbf{0.48}} & \textcolor{blue}{32.44} \\
          
\hline
ViT-L/16  & Clean & 78.83 & 100.00 \\
          & FGSM  & 48.21 & 14.44 \\
          & PGD  & 6.30 & 37.87 \\
          & AutoAttack & 6.87 & 45.58\\
          & JPEG & 68.67 & 29.27 \\
          & PCA  & 67.32 & 27.53 \\
          
          & \textcolor{blue}{JPEG $\rightarrow$ FGSM } & \textcolor{blue}{33.63} & \textcolor{blue}{31.53} \\
          & \textcolor{blue}{JPEG $\rightarrow$ PGD } & \textcolor{blue}{\textbf{0.59}} & \textcolor{blue}{30.46} \\
          
          & \textcolor{blue}{PCA $\rightarrow$ FGSM} & \textcolor{blue}{23.40} & \textcolor{blue}{30.86} \\
          & \textcolor{blue}{PCA $\rightarrow$ PGD } & \textcolor{blue}{0.60} & \textcolor{blue}{32.42} \\
\hline
\end{tabular}%
}
\end{table}

Tables~\ref{tab:cifar10}, \ref{tab:cifar100}, and \ref{tab:imagenet} summarize results. Several consistent trends emerge.

\paragraph{Compression amplifies adversarial vulnerability.}
Across datasets and architectures, applying adversarial perturbations after compression leads to significantly lower accuracy than standard pixel-space attacks at comparable PSNR levels. For example, on CIFAR-100 with ResNet-18, FGSM ($\epsilon=0.03$) reduces accuracy to 23.4\%, whereas JPEG$\rightarrow$FGSM reduces accuracy further to 8.5\% despite similar perceptual distortion. Similar amplification effects are observed for PGD and across deeper models.

\paragraph{Effect persists across model scale and dataset complexity.}
The amplification phenomenon holds for ResNet-18, ResNet-34, and ResNet-50, and is consistent on both CIFAR and ImageNet. This indicates that the effect is not architecture-specific but stems from the interaction between compression and decision geometry.

\subsection{Decision Space Reduction Analysis}
\label{sec:dsr}

\begin{figure*}[t]
\centering
\begin{minipage}[t]{0.32\textwidth}
\centering
\includegraphics[width=\linewidth]{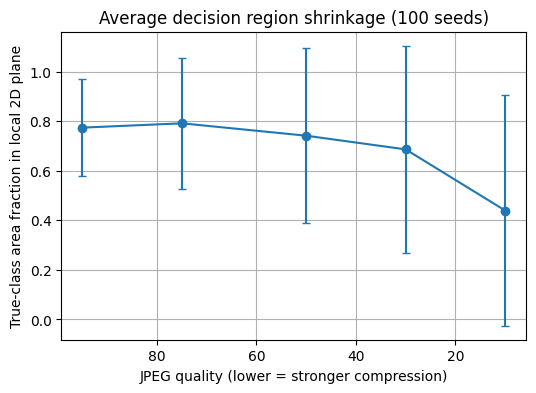}
\caption*{\small (a) True-class area fraction $A$}
\end{minipage}\hfill
\begin{minipage}[t]{0.32\textwidth}
\centering
\includegraphics[width=\linewidth]{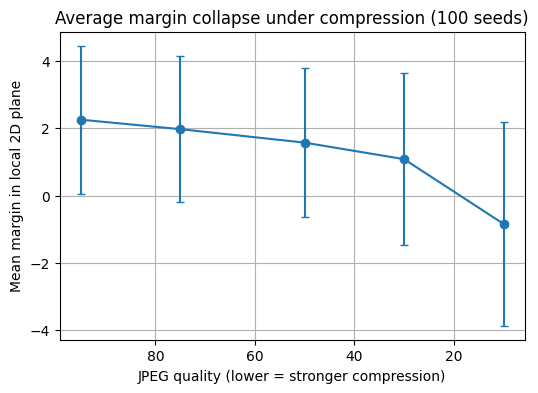}
\caption*{\small (b) Mean margin $\bar{m}$}
\end{minipage}\hfill
\begin{minipage}[t]{0.32\textwidth}
\centering
\includegraphics[width=\linewidth]{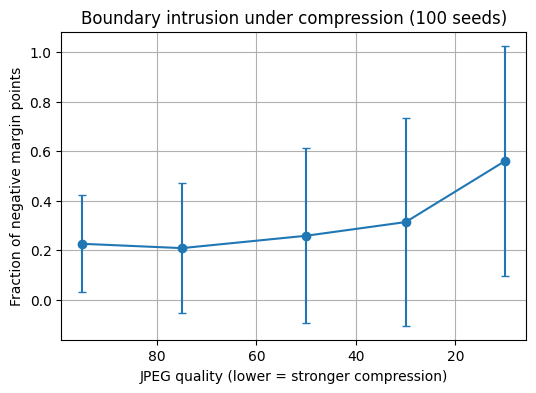}
\caption*{\small (c) Boundary intrusion $B$}
\end{minipage}

\caption{Decision space reduction under compression averaged over 100 correctly classified seeds. As JPEG quality decreases (stronger compression), (a) the fraction of the local neighborhood assigned to the true class decreases, (b) the mean margin collapses, and (c) the fraction of negative-margin points increases, indicating that decision boundaries move closer and intrude more frequently into the local neighborhood.}
\label{fig:dsr}
\end{figure*}

To explain why compression-aware perturbations are consistently more effective, we quantify how compression reshapes the \emph{local} decision geometry around clean inputs. Our goal is to measure whether compression contracts the region around an input that is assigned to its true class, and whether it simultaneously increases decision boundary proximity and boundary density.

\paragraph{Local 2D neighborhood construction.}
For each correctly classified seed $(x,y)$, we probe a two-dimensional neighborhood by sampling points on a plane through $x$:
\begin{equation}
x(\alpha,\beta) = \mathrm{clip}\big(x + \alpha u + \beta v,\, 0,\, 1\big),
\label{eq:plane}
\end{equation}
where $\alpha,\beta \in [-r,r]$, and $\mathrm{clip}(\cdot)$ enforces valid pixel bounds. We choose $u$ to be the unit-norm gradient direction of the cross-entropy loss (computed at the clean input),
\begin{equation}
u = \frac{\nabla_x \ell\big(f(x),y\big)}{\|\nabla_x \ell\big(f(x),y\big)\|_2},
\label{eq:u}
\end{equation}
and obtain $v$ by sampling a random direction and orthogonalizing it against $u$ via Gram--Schmidt, followed by $\ell_2$ normalization. This produces an informative plane that is anchored to a locally adversarial direction while still spanning a diverse neighborhood.

\paragraph{Compression-in-the-loop evaluation.}
To isolate the effect of compression on decision geometry, we evaluate predictions and margins either on the original samples $x(\alpha,\beta)$ (no compression) or after applying compression $C$ (compression-in-the-loop). For brevity, let $x_{\alpha,\beta} = x(\alpha,\beta)$. We compute logits
\begin{equation}
z_{\alpha,\beta} =
\begin{aligned}[t]
& f(x_{\alpha,\beta}) && \text{(no compression)},\\
& f(C(x_{\alpha,\beta})) && \text{(compression-in-the-loop)} ,
\end{aligned}
\label{eq:logits}
\end{equation}
and define the predicted label and true-class margin as
\begin{equation}
\hat{y}_{\alpha,\beta} = \arg\max_{k} \, z_{\alpha,\beta}^{(k)},
\label{eq:pred}
\end{equation}
\begin{equation}
m_{\alpha,\beta} = z_{\alpha,\beta}^{(y)} - \max_{k \neq y} z_{\alpha,\beta}^{(k)} .
\label{eq:margin}
\end{equation}

\paragraph{Metrics.}
Given a uniform grid $\mathcal{G}$ over $(\alpha,\beta)$, we compute three complementary proxies for decision space reduction:
\begin{align}
A(x,y) &= \frac{1}{|\mathcal{G}|}\sum_{(\alpha,\beta)\in\mathcal{G}}
\mathbb{I}\big[\hat{y}(\alpha,\beta)=y\big] \label{eq:area}\\
\bar{m}(x,y) &= \frac{1}{|\mathcal{G}|}\sum_{(\alpha,\beta)\in\mathcal{G}} m(\alpha,\beta) \label{eq:meanmargin}\\
B(x,y) &= \frac{1}{|\mathcal{G}|}\sum_{(\alpha,\beta)\in\mathcal{G}}
\mathbb{I}\big[m(\alpha,\beta) < 0\big] \label{eq:boundaryintrusion}
\end{align}
where $A(x,y)$ measures the fraction of the local neighborhood assigned to the correct class, $\bar{m}(x,y)$ measures how confidently the neighborhood supports the true class on average, and $B(x,y)$ measures how frequently the decision boundary intrudes into the neighborhood (negative margin indicates that the true class is not top-1).

\paragraph{Averaging across seeds and compression strengths.}
We compute the above metrics across $N=100$ correctly classified seeds and report the mean and standard deviation as a function of JPEG quality $q \in \{95,75,50,30,10\}$. Figure~\ref{fig:dsr}(a) shows that the true-class region fraction $A(x,y)$ decreases as compression strengthens, indicating contraction of the local decision region. Figure~\ref{fig:dsr}(b) shows a corresponding collapse in mean margin $\bar{m}(x,y)$, and Figure~\ref{fig:dsr}(c) shows that boundary intrusion $B(x,y)$ increases substantially at lower quality factors. Together, these trends provide quantitative evidence that compression brings decision boundaries closer and increases the density of competing classes in local neighborhoods.

\paragraph{Link to robustness.}
Decision space reduction offers a mechanistic explanation for the amplification observed in compression-aware attacks. When $A(x,y)$ shrinks and $\bar{m}(x,y)$ decreases, a larger fraction of perturbations with fixed magnitude will cross the decision boundary, which is reflected by an increase in $B(x,y)$. This directly supports our hypothesis that compression acts as an adversarial amplifier by contracting local margins and increasing boundary proximity.

\subsection{Attack and Then Compress}

Compression alone causes only moderate accuracy degradation, whereas applying adversarial perturbations in compressed space leads to disproportionately large performance drops. This already suggests that compression does not merely remove information but changes the local decision geometry. To further disentangle information loss from geometric effects, we also study the reverse order: first apply the adversarial attack in pixel space and then compress the perturbed image.

Importantly, this order-dependent behavior is predicted by the decision space reduction hypothesis. Compression applied to clean inputs reduces $\mathrm{Vol}(\mathcal{R}(x))$, increasing boundary proximity before any perturbation is added. In contrast, compression applied to $x+\delta$ can remove components of $\delta$ that were aligned with boundary-crossing directions, effectively increasing the margin relative to the perturbed sample. This provides a geometric explanation for why purification-style compression can restore accuracy even though compression-in-the-loop increases vulnerability.

\begin{figure}[t]
\centering
\includegraphics[width=\linewidth]{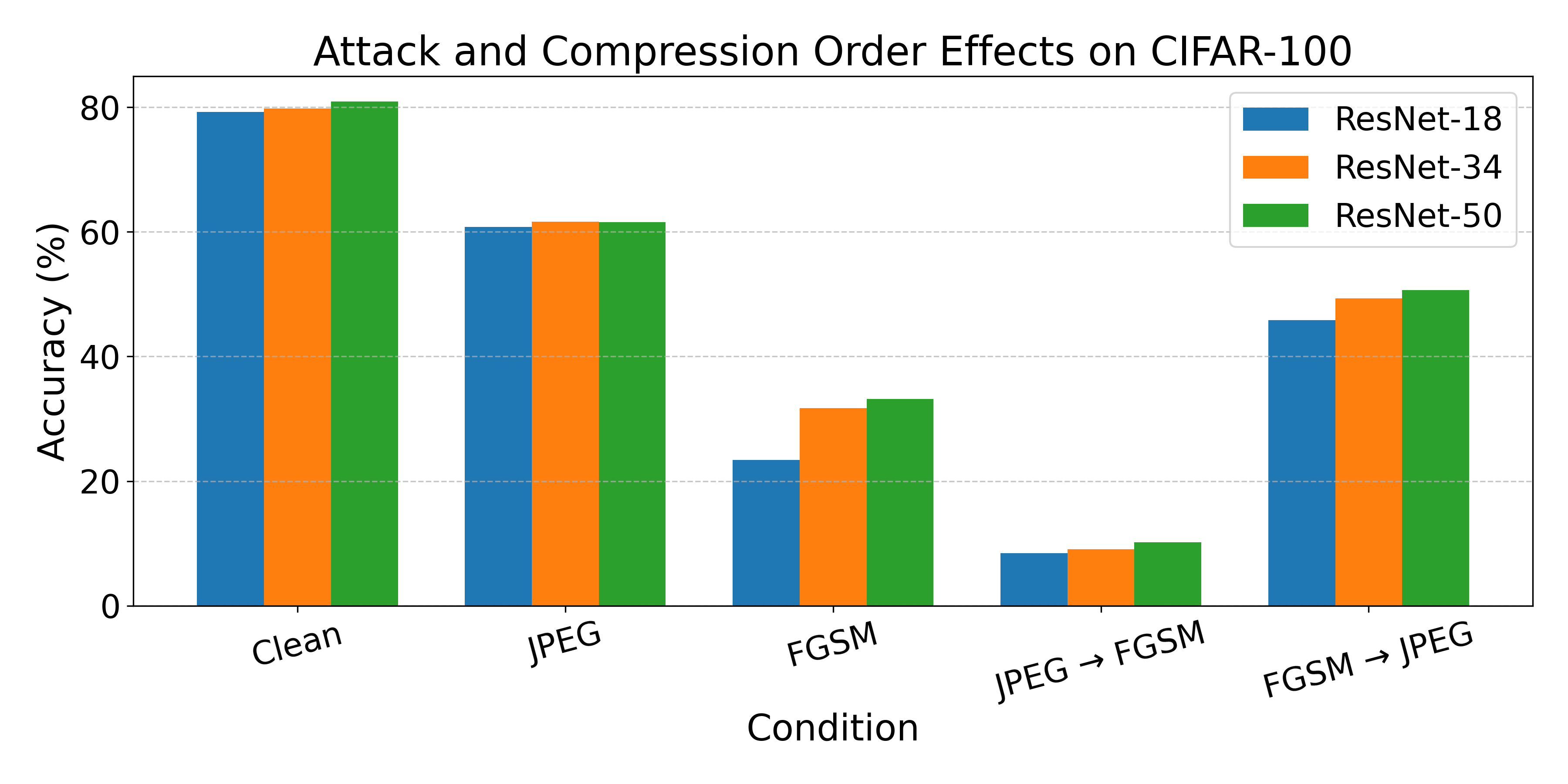}
\caption{Effect of operation order on CIFAR-100 robustness. Applying compression before the attack (JPEG$\rightarrow$FGSM) causes significantly larger accuracy drops than attacking first and then compressing (FGSM$\rightarrow$JPEG), despite similar perceptual distortion levels.}
\label{fig:attack_compress_order}
\end{figure}

The results are shown in Figure~\ref{fig:attack_compress_order}. Across all ResNet architectures, the order of operations has a substantial impact. When compression precedes the attack (JPEG$\rightarrow$FGSM), accuracy drops drastically (e.g., $79.26\%\rightarrow 8.46\%$ for ResNet-18), whereas attacking first and then compressing (FGSM$\rightarrow$JPEG) partially restores performance (e.g., $23.38\%\rightarrow 45.86\%$). Notably, JPEG preprocessing alone reduces accuracy only moderately, and FGSM$\rightarrow$JPEG achieves significantly higher accuracy than FGSM at comparable PSNR. This behavior is consistent across models.

These findings indicate that compression is not simply discarding signal but reshaping the decision space. When compression is applied before the attack, the effective decision region is already contracted, making it easier for small perturbations to cross decision boundaries. In contrast, when compression follows the attack, it can attenuate perturbation components, acting similarly to previously observed compression-based defenses. The strong asymmetry between JPEG$\rightarrow$FGSM and FGSM$\rightarrow$JPEG therefore supports our hypothesis that compression fundamentally alters local decision geometry rather.

\subsection{Ablation on $\epsilon$}

\begin{table}[t]
\centering
\caption{Effect of perturbation budget $\epsilon$ on classification accuracy (\%) under 
pixel-space (FGSM, PGD) and compression-aware attacks on CIFAR-100 with ResNet-18. Compression-aware attacks consistently outperform pixel-space counterparts across 
all budgets.}
\label{tab:epsilon_ablation}
\setlength{\tabcolsep}{5pt}
\begin{tabular}{lcccc}
\toprule
\textbf{Attack} 
    & $\epsilon=0.02$ & $\epsilon=0.04$ 
    & $\epsilon=0.06$ & $\epsilon=0.08$ \\
\midrule
FGSM              & 42.10 & 22.80 & 10.03          & 4.50 \\
JPEG$\to$FGSM     & 38.20 & 19.50 & 8.46  & 3.80 \\
PCA$\to$FGSM      & 40.10 & 21.30 & 9.77           & 4.20 \\
PatchSVD$\to$FGSM & 39.60 & 20.90 & 9.42           & 3.95 \\
\midrule
PGD               & 28.50 &  0.04 & 0.00          & 0.00 \\
JPEG$\to$PGD      & 22.10 &  0.01 & 0.00  & 0.00 \\
PCA$\to$PGD       & 24.30 &  0.00 & 0.00           & 0.00 \\
PatchSVD$\to$PGD  & 23.80 &  0.02 & 0.00          & 0.00 \\
\bottomrule
\end{tabular}
\end{table}

\paragraph{Sensitivity to perturbation budget.}
Table~\ref{tab:epsilon_ablation} reports classification accuracy on CIFAR-100 (ResNet-18)
as the perturbation budget $\epsilon$ varies from $0.02$ to $0.08$ for both pixel-space
and compression-aware attacks.
Two consistent trends emerge.
First, compression-aware attacks (JPEG$\to$FGSM, PCA$\to$FGSM, PatchSVD$\to$FGSM
and their PGD counterparts) achieve strictly lower accuracy than their pixel-space
equivalents at every budget level, confirming that the amplification effect identified
in Section~4.3 is not specific to any single $\epsilon$ choice.
Second, the accuracy gap between pixel-space and compression-aware variants
grows with $\epsilon$: at small budgets the two settings produce comparable
degradation, but as the budget increases the compression-aware attacks drive accuracy
toward zero more rapidly.

This behaviour is directly predicted by the decision space reduction hypothesis,
compression contracts the local true-class region before any perturbation is applied,
so a larger fraction of fixed-budget perturbations cross decision boundaries compared
to the uncompressed setting, and this disparity compounds as $\epsilon$ grows. Overall, these results suggest that compression not only amplifies vulnerability, but also lowers the effective perturbation budget required to achieve a given level of degradation.

\subsection{Evaluating Performance of Robust Models}

\begin{table}[t]
\centering
\caption{Compression-aware attacks on CIFAR-100 ($\ell_\infty$, $\epsilon=8/255$).
Models: DMAT, FDA and MeanSparse use WRN-70-16, MixedNuts uses ResNet-152 + WRN-70-16,
and DKLDL uses WRN-28-10. Clean and AutoAttack (AA) accuracies are from
RobustBench~\cite{croce2020robustbench}. We reimplement CoFAT~\cite{li2025towards} using WRN-70-16 and report results. Ours$_{1}$ refers to JPEG + FGSM and Ours$_{2}$ refers to JPEG + PGD.}
\label{tab:robust_models}
\setlength{\tabcolsep}{4.5pt}

\resizebox{\linewidth}{!}{
\begin{tabular}{lcccccc}
\toprule
\textbf{Model} 
    & \textbf{Clean} & \textbf{AA} 
    & \textbf{FGSM} & \textbf{PGD}
    & \textbf{Ours$_{1}$} & \textbf{Ours$_{2}$} \\
\midrule
DMAT        & 75.22 & 42.66 & 48.52 & 41.24 & 39.69 & 36.74 \\
MeanSparse  & 75.13 & 42.25 & 48.27 & 42.58 & 39.58 & 37.22 \\
MixedNuts   & 83.08 & 41.80 & 46.78 & 44.71 & 38.87 & 36.53 \\
DKLDL       & 73.85 & 39.18 & 44.52 & 41.06 & 37.75 & 35.91 \\
FDA & 63.56 &	34.64 &	38.73 & 37.79 & 31.58 & 29.82 \\
CoFAT & 75.22 & 49.72 & 55.51 & 53.35 & 47.75 & 45.95 \\
\bottomrule
\end{tabular}
}
\end{table}

A key open question is whether the amplification effect identified  persists against models that have been explicitly hardened through adversarial training.
To address this, Table~\ref{tab:robust_models} evaluates four
state-of-the-art adversarially trained models drawn from the RobustBench
leaderboard~\cite{croce2020robustbench} under both standard and
compression-aware attacks on CIFAR-100.

The selected models are DMAT~\cite{dmat},
MeanSparse~\cite{amini2024meansparse}, FDA~\cite{rebuffi2021fixing},
MixedNuts~\cite{bai2024mixednuts}, and
DKLDL~\cite{cui2024decoupled} represent the current
state of the art in $\ell_\infty$-robust classification, with
AutoAttack robust accuracies ranging from $39.18\%$ to $42.66\%$
at $\epsilon = 8/255$.
Despite their substantially improved margins relative to
standard models, compression-aware attacks (JPEG$\to$FGSM,
JPEG$\to$PGD) continue to reduce accuracy below the levels achieved
by their pixel-space counterparts, indicating that decision space
reduction operates independently of the robustness conferred by
adversarial training.

Notably, the absolute accuracy gap between pixel-space and
compression-aware attacks narrows compared to standard models,
consistent with the interpretation that adversarial training
enlarges classification margins and thereby partially counteracts
the contraction induced by compression.
Nevertheless, the residual amplification effect suggests that
compression-in-the-loop remains a meaningful threat surface
even for models optimised for adversarial robustness, and that
compression-aware adversarial training may be a necessary
direction for future work.

\section{Conclusion}

We study adversarial robustness in compression-in-the-loop settings that better reflect modern visual data pipelines, where images are routinely compressed prior to storage, transmission, or inference. Contrary to the common assumption that compression primarily acts as a defensive or purifying operation, we show that it can function as an adversarial amplifier: across datasets, architectures, and attack types, perturbations applied in compressed representations consistently induce larger performance degradation than comparable pixel-space attacks. We attribute this behavior to a geometric mechanism termed \emph{decision space reduction}, whereby compression contracts the local true-class decision region, reduces margins, and brings decision boundaries closer to inputs, as confirmed by both quantitative neighborhood analysis and decision space visualization. Our theoretical insights link this contraction to reduced effective robust radii through margin shrinkage and sensitivity amplification, while experiments on operation order reveal a strong asymmetry: compression before attack dramatically increases vulnerability, whereas compressing after attack can partially restore accuracy. Together, these findings highlight that compression fundamentally reshapes decision geometry and should be treated as part of the threat surface, underscoring the need to evaluate and design robust models under realistic, compression-aware deployment conditions. Importantly, this effect arises without requiring gradients through the compression operator, reflecting realistic threat models in which attackers operate directly on compressed inputs.

\bibliography{main}
\bibliographystyle{acm}

\end{document}